\pdfoutput=1
\documentclass[11pt]{article}

\usepackage[]{acl}

\usepackage{times}
\usepackage{latexsym}
\usepackage[T1]{fontenc}
\usepackage[utf8]{inputenc}
\usepackage{microtype}
\usepackage{inconsolata}
\usepackage{graphicx}
\usepackage{tablefootnote}

\usepackage[]{mdframed}
\usepackage{amsmath}
\usepackage{amssymb} %blackboard bold style

\title{Synthetic Data Augmentation for \\
Cross-domain Implicit Discourse Relation Recognition}

\author{Frances Yung, {\bf Varsha Suresh,}  {\bf Zaynab Reza,}\\ {\bf Mansoor Ahmad,} {\bf Vera Demberg} \\
  Saarland University, Germany \\
  \texttt{[frances, vsuresh, zreza, mahmad, vera]@lst.uni-saarland.de}}

\newcommand{\framedtext}[1]{
\par
\noindent\fbox{
    \parbox{\dimexpr\linewidth-2\fboxsep-2\fboxrule}{#1}}
    }

\begin{document}
\maketitle
\begin{abstract}
Implicit discourse relation recognition (IDRR) -- the task of identifying the implicit coherence relation between two text spans -- requires deep semantic understanding. Recent studies have shown that zero-/few-shot approaches significantly lag behind supervised models \citep{chan-etal-2024-exploring,yung-etal-2024-prompting}, but LLMs may be useful for synthetic data augmentation, where LLMs generate a second argument following a specified coherence relation. %alternative continuations of discourse in the original data, can lead to improvements.
We applied this approach in a cross-domain setting, generating discourse continuations using unlabelled target-domain data to adapt a base model which was trained on source-domain labelled data.
Evaluations conducted on a large-scale test set revealed that different variations of the approach did not result in any significant improvements. We conclude that LLMs often fail to generate useful samples for IDRR, and emphasize the importance of considering both statistical significance and comparability when evaluating IDRR models.

\end{abstract}

\section{Introduction}

IDRR is the task of identifying the covert discourse relation (DRs) between two given text spans (the arguments: \textit{Arg1} and \textit{Arg2}) in the absence of a specific discourse connective (DC) such as \textit{because} or \textit{moreover}. Explicit labelling of discourse structure is beneficial to inform LLMs in summarization tasks \cite{li-etal-2016-role,ishigaki-etal-2019-discourse,xiao-etal-2020-really,dong-etal-2021-discourse,pu-etal-2023-incorporating,pu-demberg-2024-rst}. However, IDRR is challenging both for humans \cite{hoek2021there} and models \citep[SOTA, $56.50\%$ F1 and $64.87\%$ accuracy in][]{zeng-etal-2024-global}, particularly in a cross-domain setting \cite{shi-demberg-2019-learning,atwell-etal-2022-change,liu-zeldes-2023-cant,pyatkin-etal-2023-design}. 

Prompting of large pre-trained language models (LLMs), despite the human- or even superhuman-level performance in various reasoning tasks \citep[e.g.,][]{mao2023gpteval,bang2023multitask,gilardi2023chatgpt,tornberg2023chatgpt}, was found to be not successful in IDRR. Few-shot prompting using \textit{GPT-4} only reaches $30.90\%$ F1
and $29.40\%$ accuracy on the Penn Discourse Treebank \citep[PDTB 3.0,][]{AB2/SUU9CB_2019}, and $28.87\%$ F1
and $32.67\%$ accuracy on the multi-domain DiscoGeM corpus \cite{scholman2022DiscoGeM}, even with task-specific prompt engineering \cite{chan-etal-2024-exploring,yung-etal-2024-prompting,omura-etal-2024-empirical-study}.  
On the other hand, recent work demonstrated that LLMs can instead be used to generate synthetic data to augment the PDTB~3.0, improving the performance of classes that the baseline struggles to predict \cite{omura-etal-2024-empirical-study}.

This work explores the application of synthetic data augmentation for cross-domain IDRR.  Using raw texts from different target domains, we prompt LLMs to generate discourse continuations that express specific DRs. The generated data is then used to adapt the base model trained on human-annotated data of the source domain, which is the PDTB~3.0. 
We experimented with outputs of different LLMs, prompt templates, screening strategies and various advanced methods of domain adaptation, and evaluated on the entire DiscoGeM corpus, which is $4$ times the size of the test set used in previous work \cite{omura-etal-2024-empirical-study}.
We only found marginal differences between models adapted to target-domain synthetic data compared with the base model. The benefit of generating synthetic data is unclear compared with direct application of the cross-domain base model, or using it to pseudo-label target-domain data.

We have derived the following insights from the experimental results:
\begin{enumerate}
    \item Synthetic data augmentation by LLMs does not improve IDRR under a cross-domain setting, where annotated target-domain data is unavailable for training nor validation. The synthetic data quality relies on the screening by the base model trained on large-scale annotated data from the source domain.
    \item Previous positive results in in-domain IDRR may have been overly optimistic, as improvements were observed only under specific configurations, and the results exhibited high variability. 
    
    \item Fine-grained discourse inference remains a significant challenge for LLMs, compared with NLU tasks such as binary sentiment or topic classification \cite{ubani2023zeroshotdataaug,piedboeuf-langlais-2025-evaluation}. 
    \item Manual analysis shows that the valid generated samples are often highly prototypical examples of the DR class. In contrast, DRs in real texts tend to be more ambiguous and rely on indirect inference.

\end{enumerate}

\section{Related Work}

\subsection{Synthetic data generated by LLMs}
Supervised learning algorithms rely on labeled data as training objectives but manual annotation is time- and cost-intensive. Using controlled text generation \cite{hu2017toward} with LLMs through instructional prompts, large-scale training data was created for a range of NLP tasks, such as question-answering \cite{puri-etal-2020-training},  textual similarity identification \cite{schick-schutze-2021-generating}, NLU \cite{meng2022generating,liu-etal-2022-wanli} commonsense reasoning \cite{yang-etal-2020-generative}, and dialogue classification \cite{sharma2023team}, etc.

Recent findings indicate that while some tasks benefit from synthetic augmentation, others do not \cite{ubani2023zeroshotdataaug,moller-etal-2024-parrot,piedboeuf-langlais-2025-evaluation}. For example, synthetic data augmentation has been shown to significantly improve performance in movie review classification (positive vs. negative) and question type identification (e.g., whether a question seeks a "reason" or a "number"). However, in ambiguous tasks like irony detection,
models trained with synthetic data often underperform the baseline.

\subsection{Implicit DR recognition}

The current SOTA of IDRR models are mostly based on fine-tuning or prompt-tuning of the RoBERTa model \cite{xiang-etal-2022-connprompt,zhou-etal-2022-prompt-based,zhao-etal-2023-infusing, jiang-etal-2023-global,zeng-etal-2024-global}. 
Cross-domain IDRR remains challenging and understudied. We started our experiments with GOLF \cite{jiang-etal-2023-global}, one of the SOTA models trained on PDTB, but found no significant improvement over a standard RoBERTa model.
Several previous studies have found that models do not generalize well to out-of-domain data \cite{shi-demberg-2019-learning,atwell-etal-2021-discourse,liu-etal-2021-dmrst,scholman-etal-2021-comparison,kurfali-ostling-2021-lets,liu-zeldes-2023-cant,braud-etal-2023-disrpt,li-etal-2024-discourse}. 

Furthermore, attempts to classify implicit DR by zero/few-shot prompting were not fruitful -- both a standard multiple-choice template \cite{chan-etal-2024-exploring} and multi-step templates with verification questions \cite{yung-etal-2024-prompting} result in performance significantly below that of supervised models.

\citet{omura-etal-2024-empirical-study} introduced an approach to augment the PDTB with synthetic data. Specifically, given an \textit{Arg1} from the original PDTB training set and a DR label, an LLM is prompted to generate an alternative \textit{Arg2}. The generated samples undergo a secondary filtering stage via few-shot prompting to discard ambiguous cases. During model training, a weighted loss is applied to balance the original and synthetic samples.
The augmentation strategy targets the most confusing DR classes, i.e. those with the lowest recall on the PDTB validation set. Performance gains were observed when augmenting the top-3 most confusing classes, but not top-1 nor top-5.\footnote{for the RoBERTa$_{base}$ classifier. When using RoBERTa$_{large}$, augmentation of the top-5 classes was beneficial.}

Compared to a standard RoBERTa$_{base}$ model \cite{DBLP:journals/corr/abs-1907-11692}, the reported improvements were modest: accuracy increased from $64.2$ to $64.8$, and macro-F1 improved from $57.1$ to $59.5$. However, results varied across the $3$ runs, with fluctuations ranging from $\pm0.4$ to $\pm1.6$ points.

\section{Experiment}
\subsection{Data}
We set out to use synthetic discourse samples to adapt an IDRR model trained on source-domain data, which is the PDTB~3.0 \cite{AB2/SUU9CB_2019}, to predict implicit DRs in the target domains, which are the sub-corpora in DiscoGeM~1.5 \cite{scholman2022DiscoGeM,yung-demberg-2025-crowdsourcing}. The target domains include \textit{Europarl} (EP), \textit{Wikipedia} (WK) and \textit{novel} (NV). We experiment based on a real-life scenario, where labelled target domain data is \textit{unavailable}, i.e. the DiscoGeM data is used only for testing. However, it is assumed that the target domain is known. 
The label distribution of the data we used and more details about the data can be found in Section \ref{sec:data} in the Appendix.

For the generation of the synthetic data, we collected raw texts from similar sources as the target domains, EP texts from the Europarl Direct Corpus \cite{Koehn-Europarl-2005,Cartoni-Directional-2012}; WK texts from the Wikipedia;
and NV texts from the Opus Book Corpus \cite{tiedemann-2012-parallel}, omitting proceedings, articles or novels contained in DG.

\subsection{Methodology}
\begin{table*}[h]
    \centering \small
    \begin{tabular}{@{}ll@{}}
        \hline
        \textbf{LLMs} & \textit{Mistral-7B-Instruct}, \textit{Llama3.1 8B-Instruct}, \textit{gemma2-9B} \\ 
        \hline
        \multicolumn{2}{@{}l}{\textbf{prompt template} (more details in Appendix \ref{sec:prompt})}\\
        %\hline
        DC-prompt & lexicalize the DR to be generated by a connective (DC), e.g. \textit{because} for the \textsc{causal} relation \\
        DR-prompt & directly prompt by the DR label, providing the definition on the annotation manual\\
        \hline
        
        \multicolumn{2}{@{}l}{\textbf{screening method} (more details in Appendix~\ref{sec:confusion screen})}\\
        
        strict screen & only include samples where the intended DR matches the prediction by the base model\\
        confusion screen & exclude samples where the predicted label is a frequent misprediction of the intended label \\
        combi screen & combination: apply the confusion screen if the intended DR is rare, otherwise the strict screen\footnotemark{}\\
        
        \hline
        
        \multicolumn{2}{@{}l}{\textbf{adaptation model}}\\
       
         PDTB $+$ DG$_{\text{syn}}$ 
         & a RoBERTa$_{base}$ model trained on a direct combination of the PDTB and synthetic data\\
         
         PDTB $\rightarrow$ DG$_{\text{syn}}$
         & the PDTB-trained RoBERTa$_{base}$ model adapted to the synthetic data by prefix-tuning\\
         PDTB $\rightarrow_{IV}$ DG$_{\text{syn}}$
         & include an invariance loss \cite{zhou2020deep,tzeng2014deep} alongside the standard cross-entropy\\ 
         &  loss to encourage the model to learn features that are indistinguishable between real and synthetic data\footnotemark{}\\
         
         \multicolumn{2}{@{}l}{\textbf{target-domain data configuration}}\\
        
         domain-specific & one model specifically adapted to the synthetic data of each domain\\
         domain-mixed & a single model trained on the combined synthetic data of all domains, with a domain token
        
         prepended to \\
         &the beginning of each sample \cite{yung-etal-2022-label}\\
         \hline
    \end{tabular}
    \caption{Variants of the synthetic DR augmentation approach explored in the experiment.}
    \label{tab:variations}
\end{table*}
Table~\ref{tab:variations} summarizes the different methodological variants we explored.
To harness the LLM's strength of left-to-right generation, we prompt the LLM to produce the continuation of a discourse prefix as in \citet{omura-etal-2024-empirical-study}. Given a sentence (the \textit{Arg1}) from the raw text of the target domain and a DR label, an LLM is prompted to generate the following sentence (the \textit{Arg2}). One in-context example selected from DiscoGeM is provided. We experimented with two different prompt templates.
The synthetic data was generated using three LLMs, including: \textit{Mistral-7B-Instruct-v 0.2} \cite{jiang2023mistral}, \textit{Llama3.1 8B-Instruct} \cite{dubey2024llama}, and \textit{gemma2 9B} \cite{team2024gemma}.

The generated DRs then undergo a selection process to remove noisy instances. 
  
Since the zero-shot performance of LLMs significantly lags behind supervised models \cite{yung-etal-2024-prompting}, we trained a RoBERTa$_{base}$ model \cite{liu2019roberta} on the PDTB~3.0 to predict the DR of the synthetic samples. The synthetic samples were selected based on the predictions of this source-domain model. We compared three screening strategies.  

The screened synthetic DR instances (statistics in Table~\ref{tab:generation} in the Appendix) are then used as domain-specific data to adapt the source-domain model.  Similarly, we evaluated several methods and configurations, such as prefix tuning \cite{li2021prefix} vs.~simple data concatenation.

We compare the proposed models with the baseline model trained on PDTB (i.e.~the model for filtering), the SOTA GOLF model \cite{jiang-etal-2023-global} trained on PDTB, and pseudo-labeling \cite{yarowsky1995unsupervised}. The pseudo-labelled data (DG$_{pseudo}$) are produced by using the baseline model to label adjacent sentence pairs in target-domain raw data.

For each of the $14$ DR labels in PDTB 3.0, we generated one synthetic sample using each LLM, based on $4,000$ randomly sampled raw sentences in each target domain.

A total of $12,000$ pseudo-labelled instances per domain were used for comparison, roughly matching the size of the screened synthetic data (see Table~\ref{tab:generation}). The combined data in the domain-mixed configuration was also downsampled to approximately $10,000$ instances, while preserving the per-domain and per-class distributions to ensure comparability.

All domain adaptation models are trained for $3$ epochs with a learning rate of $1e-04$, primarily chosen to prevent overfitting \cite{yang2024can}.

We set the embedding dimension of the prefix-tuning parameters to $512$ leading to $\approx$ 7M trainable parameters compared to $\approx$ 130M parameters required for full fine-tuning.

\subsection{Results}
\begin{table*}[h]
    \centering \small
    \begin{tabular}{@{}l@{}|@{}l@{}ll|@{}ll@{}|ll|ll|ll@{}}
        \hline
         Model &LLM & tpl.& screen 
         &\multicolumn{2}{c|}{tgt. domain data}
         & \multicolumn{2}{c|}{EP} 
         & \multicolumn{2}{c|}{WK}
         & \multicolumn{2}{c}{NV}\\
         \cline{5-12}
         &&&&config.& size\footnotemark{}
         
         & F1 & Acc & F1 & Acc & F1 & Acc \\
         \hline\hline
         
         Baseline PDTB 
         &-&-&-&-& 0 &
         $21.03$&$42.00$&$22.81$&$45.58$&$21.94$&$43.98$\\
         GOLF PDTB
         &-&-&-&-& 0 &
         $21.30$&$42.05$&$23.98$&$46.29$&$21.20$&$42.93$\\
         \hline
         PDTB $\rightarrow$  DG$_{\text{syn}}$
         &llama3 & DC & strict & specific 
         &$8680$&$21.69$&$41.74$&$22.33$&$47.32$&$22.89$&$44.19$ \\
         & gemma2\hspace{0.1cm} & DC & strict & specific
         &$8315$&$21.94$&$41.88$&$23.99$&$46.67$&$23.12$&$44.98$ \\
         
         & mistral & DC & strict & specific
         &$10546$&$21.47$& $40.54$&$24.42$&$47.05$&$22.68$&$44.88$ \\
         \cline{2-12}
         & mistral & DC & confuse & specific
         &$43214$&$11.90^*$&$21.73^*$&$16.94^*$&$35.07^*$&$15.47^*$&$31.54^*$ \\
         & mistral & DC & combi & specific
         &$18286$&$16.72^*$&$32.47^*$&$19.41^*$&$41.90^*$&$18.73^*$&$39.05^*$ \\
          \cline{2-12}
         & mistral & DR & strict & specific
         &$12376$&$21.62$&$42.16$&\textbf{24.87}&$47.59^*$&$22.86$&\textbf{46.67} \\
          & mistral & DR & strict & mixed
         &$10441$&$21.58$&$41.77$&$24.03$&$47.21$&\textbf{23.19}&$45.54$ \\
         & mistral & DC & strict & mixed
         &$10441$&\textbf{22.40}$^*$&$42.05$&$23.73$&$46.78$&$22.50$&$44.19$ \\
         \hline
         PDTB $+$  DG$_{\text{syn}}$  
         & mistral & DC & strict & specific
         &$12356$&$21.32$&$39.72^*$&$23.82$&$46.94$&$22.80$&$44.86$ \\
         PDTB $\rightarrow_{IV}$  DG$_{\text{syn}}$
         & mistral & DC & strict & specific
         &$12376$&$21.57$&$40.32$&$24.12$&$47.53$&$22.41$&$44.60$ \\
         
         & mistral & DC & strict & mixed
         &$10441$&$22.03^*$&$41.46$&$22.38$&$47.05$&$22.50$&$43.74$ \\
         \hline
         PDTB $+$  DG$_{\text{pseudo}}$ 
         & - & - & - & specific
         &$12000$&$20.81$&$41.28$&$23.63$&$46.29$&$23.07^*$&$44.12$ \\ 
         PDTB $\rightarrow$  DG$_{\text{pseudo}}$ 
         & - & - & - & specific
         &$12000$&$20.71$&\textbf{42.37}&$23.42$&\textbf{47.64}$^*$&$22.30$&$44.29$ \\ 
         & -& -&-  & mixed
         &$10000$&$21.78^*$&$42.01$&$23.90$&$46.83$&$21.80$&$43.00$ \\
         PDTB $\rightarrow_{IV}$  DG$_{\text{pse}}$ 
         & - & - & - & specific
         &$12000$&$21.01$&$41.88$&$24.37$&$47.21$&$21.66$&$43.39$ \\ 
         \hline
         \hline
\end{tabular}
    \caption{Performance of model variants evaluated on the sub-corpora of DiscoGeM. The best scores are bolded. Significant differences from the baseline model, based on variations across runs, are marked with $*$.} \label{tab:results}
\end{table*}
Table \ref{tab:results} presents the major comparison of the model variants. We focused on models using generations from \textit{Mistral}, as it had the highest screening pass rate, indicating superior generation quality. All results are averaged values based on $3$ random seeds. 
For evaluation, we computed accuracy and classwise F1 in line with previous works \citep[e.g.][]{xue-etal-2015-conll}, where macro-F1 scores are averaged across all classes occurring in the test set. For items with multiple gold labels, predictions matching any of the gold labels are considered correct, and the unmatched alternative labels are excluded from the classwise F1 calculation.
We assessed the statistical significance of the difference between each model and the baseline using t-tests conducted over the results from the $3$ experimental runs.

It can be seen in Table~\ref{tab:results} that none of the model variants consistently outperform the baseline across domains and evaluation metrics. Considering the variation across runs, most results do not show statistically significant differences from the baseline. This suggests that numerical differences, up to $2.7\%$ points, are primarily due to network randomness rather than genuine performance improvements. 
%The SOTA GOLF model also does not outperform the baseline on DiscoGeM, highlighting the challenge of cross-domain IDRR. 
No clear advantage is observed over the more straightforward pseudo-labeling method.
The only consistent and significant observation is the under-performance of the models using the \textit{confusion} and \textit{combi} screens. This proves that the synthetic samples, without strict guidance by a supervised model,
are not helpful. 

\section{Discussion and conclusion}%
Contrary to the improvements reported in previous studies, our results did not confirm the benefits of synthetic data augmentation for IDRR in a cross-domain setting. This is in line with recent reports that synthetic samples generated by LLMs do not improve abstract and ambiguous tasks, such as irony detection \cite{piedboeuf-langlais-2025-evaluation}.
The quality of synthetic data is crucial, and given that it is not sufficient per se, it further depends on the filtering model trained on annotated data. 

We found that the performance of IDRR models exhibits high variance, as also shown in previous studies. The conventional evaluation protocol using macro F1 may not be optimal for IDRR, as the F1 scores of rare classes are unlikely to be significant due to the skewed data distribution.  The evaluation methods for instances with multiple gold labels also vary across studies, leading to inconsistencies.\footnotemark

Manual inspection of the synthetic data shows that the generated DRs are less ambiguous than real data, as instructing the LLMs to generate specific DR samples implicitly biases them to produce prototypical examples. Real DRs are often ambiguous, lacking clear-cut cues (examples in Appendix  \ref{sec:examples}).

\addtocounter{footnote}{-4} 
\stepcounter{footnote}\footnotetext{defined as inter-sentential implicit DR with $5\%$ or more distribution in PDTB 3.0.}
\stepcounter{footnote}\footnotetext{The final loss is defined as $L_{\text{CE}} - \lambda L_{\text{IV}}$, where $\lambda$ is set to $0.1$ based on a search from $\{0.1, 0.3, 0.5\}$.}
\stepcounter{footnote}\footnotetext{The synthetic data sizes of the domain-specific models are averaged across the models of the three domains.}
\stepcounter{footnote}\footnotetext{While some works, including ours, discard unmatched alternative gold labels; other works, such as \citet{omura-etal-2024-empirical-study}, count them as true positives in the F1 calculation.} 

Finally, the datasets chosen for domain adaptation, PDTB and DG, were annotated by trained professionals and crowd workers respectively. On top of the domain shift, the annotators in these two datasets may not have identified DR labels in the same way. Annotated data from DG would be necessary to effectively guide the adaptation process.

\section{Limitation}
We did not extensively refine the prompt templates used for generating DR samples in order to generate discourses that are closer to natural examples. As discussed earlier, the generated samples were less ambiguous than naturally occurring discourses. A possible improvement could involve explicitly instructing LLMs to generate ambiguous examples or instances with multiple DR interpretations. However, pseudo-labeling remains a more promising approach for capturing DR ambiguity, as it leverages real texts rather than synthetic ones.
Another limitation of our study is the limited number of experimental runs due to time constraints. Future work could explore additional factors, such as the effect of synthetic sample size and the impact of contrastive generation, such as pairing different Arg1s in each sample v.s. using a single Arg1 with multiple continuations representing different DR senses.

\bibliography{custom}

\begin{thebibliography}{59}
\providecommand{\natexlab}[1]{#1}

\bibitem[{Atwell et~al.(2021)Atwell, Li, and Alikhani}]{atwell-etal-2021-discourse}
Katherine Atwell, Junyi~Jessy Li, and Malihe Alikhani. 2021.
\newblock \href {https://doi.org/10.18653/v1/2021.sigdial-1.34} {Where are we in discourse relation recognition?}
\newblock In \emph{Proceedings of the 22nd Annual Meeting of the Special Interest Group on Discourse and Dialogue}, pages 314--325, Singapore and Online. Association for Computational Linguistics.

\bibitem[{Atwell et~al.(2022)Atwell, Sicilia, Hwang, and Alikhani}]{atwell-etal-2022-change}
Katherine Atwell, Anthony Sicilia, Seong~Jae Hwang, and Malihe Alikhani. 2022.
\newblock \href {https://doi.org/10.18653/v1/2022.findings-acl.68} {The change that matters in discourse parsing: Estimating the impact of domain shift on parser error}.
\newblock In \emph{Findings of the Association for Computational Linguistics: ACL 2022}, pages 824--845, Dublin, Ireland. Association for Computational Linguistics.

\bibitem[{Bang et~al.(2023)Bang, Cahyawijaya, Lee, Dai, Su, Wilie, Lovenia, Ji, Yu, Chung et~al.}]{bang2023multitask}
Yejin Bang, Samuel Cahyawijaya, Nayeon Lee, Wenliang Dai, Dan Su, Bryan Wilie, Holy Lovenia, Ziwei Ji, Tiezheng Yu, Willy Chung, et~al. 2023.
\newblock A multitask, multilingual, multimodal evaluation of chatgpt on reasoning, hallucination, and interactivity.
\newblock \emph{arXiv preprint arXiv:2302.04023}.

\bibitem[{Braud et~al.(2023)Braud, Liu, Metheniti, Muller, Rivi{\`e}re, Rutherford, and Zeldes}]{braud-etal-2023-disrpt}
Chlo{\'e} Braud, Yang~Janet Liu, Eleni Metheniti, Philippe Muller, Laura Rivi{\`e}re, Attapol Rutherford, and Amir Zeldes. 2023.
\newblock \href {https://doi.org/10.18653/v1/2023.disrpt-1.1} {The {DISRPT} 2023 shared task on elementary discourse unit segmentation, connective detection, and relation classification}.
\newblock In \emph{Proceedings of the 3rd Shared Task on Discourse Relation Parsing and Treebanking (DISRPT 2023)}, pages 1--21, Toronto, Canada. The Association for Computational Linguistics.

\bibitem[{Cartoni and Meyer(2012)}]{Cartoni-Directional-2012}
Bruno Cartoni and Thomas Meyer. 2012.
\newblock Extracting directional and comparable corpora from a multilingual corpus for translation studies.
\newblock In \emph{Proceedings 8th International Conference on Language Resources and Evaluation (LREC)}, Istanbul, Turkey.

\bibitem[{Chan et~al.(2024)Chan, Jiayang, Wang, Jiang, Fang, Liu, and Song}]{chan-etal-2024-exploring}
Chunkit Chan, Cheng Jiayang, Weiqi Wang, Yuxin Jiang, Tianqing Fang, Xin Liu, and Yangqiu Song. 2024.
\newblock \href {https://aclanthology.org/2024.findings-eacl.47} {Exploring the potential of {C}hat{GPT} on sentence level relations: A focus on temporal, causal, and discourse relations}.
\newblock In \emph{Findings of the Association for Computational Linguistics: EACL 2024}, pages 684--721, St. Julian{'}s, Malta. Association for Computational Linguistics.

\bibitem[{Dong et~al.(2021)Dong, Mircea, and Cheung}]{dong-etal-2021-discourse}
Yue Dong, Andrei Mircea, and Jackie Chi~Kit Cheung. 2021.
\newblock \href {https://doi.org/10.18653/v1/2021.eacl-main.93} {Discourse-aware unsupervised summarization for long scientific documents}.
\newblock In \emph{Proceedings of the 16th Conference of the European Chapter of the Association for Computational Linguistics: Main Volume}, pages 1089--1102, Online. Association for Computational Linguistics.

\bibitem[{Dubey et~al.(2024)Dubey, Jauhri, Pandey, Kadian, Al-Dahle, Letman, Mathur, Schelten, Yang, Fan et~al.}]{dubey2024llama}
Abhimanyu Dubey, Abhinav Jauhri, Abhinav Pandey, Abhishek Kadian, Ahmad Al-Dahle, Aiesha Letman, Akhil Mathur, Alan Schelten, Amy Yang, Angela Fan, et~al. 2024.
\newblock The llama 3 herd of models.
\newblock \emph{arXiv preprint arXiv:2407.21783}.

\bibitem[{Gilardi et~al.(2023)Gilardi, Alizadeh, and Kubli}]{gilardi2023chatgpt}
Fabrizio Gilardi, Meysam Alizadeh, and Ma{\"e}l Kubli. 2023.
\newblock Chatgpt outperforms crowd-workers for text-annotation tasks.
\newblock \emph{arXiv preprint arXiv:2303.15056}.

\bibitem[{Hoek et~al.(2021)Hoek, Scholman, and Sanders}]{hoek2021there}
Jet Hoek, Merel~CJ Scholman, and Ted~JM Sanders. 2021.
\newblock Is there less annotator agreement when the discourse relation is underspecified?
\newblock In \emph{Proceedings of the First Workshop on Integrating Perspectives on Discourse Annotation}, pages 1--6.

\bibitem[{Hu et~al.(2017)Hu, Yang, Liang, Salakhutdinov, and Xing}]{hu2017toward}
Zhiting Hu, Zichao Yang, Xiaodan Liang, Ruslan Salakhutdinov, and Eric~P Xing. 2017.
\newblock Toward controlled generation of text.
\newblock In \emph{International conference on machine learning}, pages 1587--1596. PMLR.

\bibitem[{Ishigaki et~al.(2019)Ishigaki, Kamigaito, Takamura, and Okumura}]{ishigaki-etal-2019-discourse}
Tatsuya Ishigaki, Hidetaka Kamigaito, Hiroya Takamura, and Manabu Okumura. 2019.
\newblock \href {https://doi.org/10.26615/978-954-452-056-4_059} {Discourse-aware hierarchical attention network for extractive single-document summarization}.
\newblock In \emph{Proceedings of the International Conference on Recent Advances in Natural Language Processing (RANLP 2019)}, pages 497--506, Varna, Bulgaria. INCOMA Ltd.

\bibitem[{Ji et~al.(2015)Ji, Zhang, and Eisenstein}]{ji-etal-2015-closing}
Yangfeng Ji, Gongbo Zhang, and Jacob Eisenstein. 2015.
\newblock \href {https://doi.org/10.18653/v1/D15-1264} {Closing the gap: Domain adaptation from explicit to implicit discourse relations}.
\newblock In \emph{Proceedings of the 2015 Conference on Empirical Methods in Natural Language Processing}, pages 2219--2224, Lisbon, Portugal. Association for Computational Linguistics.

\bibitem[{Jiang et~al.(2023{\natexlab{a}})Jiang, Sablayrolles, Mensch, Bamford, Chaplot, Casas, Bressand, Lengyel, Lample, Saulnier et~al.}]{jiang2023mistral}
Albert~Q Jiang, Alexandre Sablayrolles, Arthur Mensch, Chris Bamford, Devendra~Singh Chaplot, Diego de~las Casas, Florian Bressand, Gianna Lengyel, Guillaume Lample, Lucile Saulnier, et~al. 2023{\natexlab{a}}.
\newblock Mistral 7b.
\newblock \emph{arXiv preprint arXiv:2310.06825}.

\bibitem[{Jiang et~al.(2023{\natexlab{b}})Jiang, Zhang, and Wang}]{jiang-etal-2023-global}
Yuxin Jiang, Linhan Zhang, and Wei Wang. 2023{\natexlab{b}}.
\newblock \href {https://doi.org/10.18653/v1/2023.findings-acl.510} {Global and local hierarchy-aware contrastive framework for implicit discourse relation recognition}.
\newblock In \emph{Findings of the Association for Computational Linguistics: ACL 2023}, pages 8048--8064, Toronto, Canada. Association for Computational Linguistics.

\bibitem[{Kim et~al.(2020)Kim, Feng, Gunasekara, and Lastras}]{kim-etal-2020-implicit}
Najoung Kim, Song Feng, Chulaka Gunasekara, and Luis Lastras. 2020.
\newblock \href {https://doi.org/10.18653/v1/2020.acl-main.480} {Implicit discourse relation classification: We need to talk about evaluation}.
\newblock In \emph{Proceedings of the 58th Annual Meeting of the Association for Computational Linguistics}, pages 5404--5414, Online. Association for Computational Linguistics.

\bibitem[{Koehn(2005)}]{Koehn-Europarl-2005}
Philipp Koehn. 2005.
\newblock Europarl: A parallel corpus for statistical machine translation.
\newblock In \emph{Proceedings of MT Summit X}, pages 79--86, Phuket, Thailand.

\bibitem[{Kurfal{\i} and {\"O}stling(2021)}]{kurfali-ostling-2021-lets}
Murathan Kurfal{\i} and Robert {\"O}stling. 2021.
\newblock \href {https://doi.org/10.18653/v1/2021.unimplicit-1.1} {Let{'}s be explicit about that: Distant supervision for implicit discourse relation classification via connective prediction}.
\newblock In \emph{Proceedings of the 1st Workshop on Understanding Implicit and Underspecified Language}, pages 1--10, Online. Association for Computational Linguistics.

\bibitem[{Li et~al.(2024)Li, Braud, Amblard, and Carenini}]{li-etal-2024-discourse}
Chuyuan Li, Chlo{\'e} Braud, Maxime Amblard, and Giuseppe Carenini. 2024.
\newblock \href {https://aclanthology.org/2024.codi-1.15} {Discourse relation prediction and discourse parsing in dialogues with minimal supervision}.
\newblock In \emph{Proceedings of the 5th Workshop on Computational Approaches to Discourse (CODI 2024)}, pages 161--176, St. Julians, Malta. Association for Computational Linguistics.

\bibitem[{Li et~al.(2016)Li, Thadani, and Stent}]{li-etal-2016-role}
Junyi~Jessy Li, Kapil Thadani, and Amanda Stent. 2016.
\newblock \href {https://doi.org/10.18653/v1/W16-3617} {The role of discourse units in near-extractive summarization}.
\newblock In \emph{Proceedings of the 17th Annual Meeting of the Special Interest Group on Discourse and Dialogue}, pages 137--147, Los Angeles. Association for Computational Linguistics.

\bibitem[{Li and Liang(2021)}]{li2021prefix}
Xiang~Lisa Li and Percy Liang. 2021.
\newblock Prefix-tuning: Optimizing continuous prompts for generation.
\newblock In \emph{Proceedings of the 59th Annual Meeting of the Association for Computational Linguistics and the 11th International Joint Conference on Natural Language Processing (Volume 1: Long Papers)}, pages 4582--4597.

\bibitem[{Liu et~al.(2022)Liu, Swayamdipta, Smith, and Choi}]{liu-etal-2022-wanli}
Alisa Liu, Swabha Swayamdipta, Noah~A. Smith, and Yejin Choi. 2022.
\newblock \href {https://doi.org/10.18653/v1/2022.findings-emnlp.508} {{WANLI}: Worker and {AI} collaboration for natural language inference dataset creation}.
\newblock In \emph{Findings of the Association for Computational Linguistics: EMNLP 2022}, pages 6826--6847, Abu Dhabi, United Arab Emirates. Association for Computational Linguistics.

\bibitem[{Liu and Demberg(2024)}]{pu-demberg-2024-rst}
Dongqi Liu and Vera Demberg. 2024.
\newblock \href {https://doi.org/10.18653/v1/2024.naacl-long.121} {{RST}-{L}o{RA}: A discourse-aware low-rank adaptation for long document abstractive summarization}.
\newblock In \emph{Proceedings of the 2024 Conference of the North American Chapter of the Association for Computational Linguistics: Human Language Technologies (Volume 1: Long Papers)}, pages 2200--2220, Mexico City, Mexico. Association for Computational Linguistics.

\bibitem[{Liu et~al.(2023)Liu, Wang, and Demberg}]{pu-etal-2023-incorporating}
Dongqi Liu, Yifan Wang, and Vera Demberg. 2023.
\newblock \href {https://doi.org/10.18653/v1/2023.acl-long.306} {Incorporating distributions of discourse structure for long document abstractive summarization}.
\newblock In \emph{Proceedings of the 61st Annual Meeting of the Association for Computational Linguistics (Volume 1: Long Papers)}, pages 5574--5590, Toronto, Canada. Association for Computational Linguistics.

\bibitem[{Liu and Zeldes(2023)}]{liu-zeldes-2023-cant}
Yang~Janet Liu and Amir Zeldes. 2023.
\newblock \href {https://doi.org/10.18653/v1/2023.eacl-main.227} {Why can{'}t discourse parsing generalize? a thorough investigation of the impact of data diversity}.
\newblock In \emph{Proceedings of the 17th Conference of the European Chapter of the Association for Computational Linguistics}, pages 3112--3130, Dubrovnik, Croatia. Association for Computational Linguistics.

\bibitem[{Liu(2019)}]{liu2019roberta}
Yinhan Liu. 2019.
\newblock Roberta: A robustly optimized bert pretraining approach.
\newblock \emph{arXiv preprint arXiv:1907.11692}.

\bibitem[{Liu et~al.(2019)Liu, Ott, Goyal, Du, Joshi, Chen, Levy, Lewis, Zettlemoyer, and Stoyanov}]{DBLP:journals/corr/abs-1907-11692}
Yinhan Liu, Myle Ott, Naman Goyal, Jingfei Du, Mandar Joshi, Danqi Chen, Omer Levy, Mike Lewis, Luke Zettlemoyer, and Veselin Stoyanov. 2019.
\newblock \href {https://arxiv.org/abs/1907.11692} {Roberta: {A} robustly optimized {BERT} pretraining approach}.
\newblock \emph{CoRR}, abs/1907.11692.

\bibitem[{Liu et~al.(2021)Liu, Shi, and Chen}]{liu-etal-2021-dmrst}
Zhengyuan Liu, Ke~Shi, and Nancy Chen. 2021.
\newblock \href {https://doi.org/10.18653/v1/2021.codi-main.15} {{DMRST}: A joint framework for document-level multilingual {RST} discourse segmentation and parsing}.
\newblock In \emph{Proceedings of the 2nd Workshop on Computational Approaches to Discourse}, pages 154--164, Punta Cana, Dominican Republic and Online. Association for Computational Linguistics.

\bibitem[{Mao et~al.(2023)Mao, Chen, Zhang, Guerin, and Cambria}]{mao2023gpteval}
Rui Mao, Guanyi Chen, Xulang Zhang, Frank Guerin, and Erik Cambria. 2023.
\newblock Gpteval: A survey on assessments of chatgpt and gpt-4.
\newblock \emph{arXiv preprint arXiv:2308.12488}.

\bibitem[{Meng et~al.(2022)Meng, Huang, Zhang, and Han}]{meng2022generating}
Yu~Meng, Jiaxin Huang, Yu~Zhang, and Jiawei Han. 2022.
\newblock Generating training data with language models: Towards zero-shot language understanding.
\newblock \emph{Advances in Neural Information Processing Systems}, 35:462--477.

\bibitem[{M{\o}ller et~al.(2024)M{\o}ller, Pera, Dalsgaard, and Aiello}]{moller-etal-2024-parrot}
Anders~Giovanni M{\o}ller, Arianna Pera, Jacob Dalsgaard, and Luca Aiello. 2024.
\newblock \href {https://aclanthology.org/2024.eacl-short.17/} {The parrot dilemma: Human-labeled vs. {LLM}-augmented data in classification tasks}.
\newblock In \emph{Proceedings of the 18th Conference of the European Chapter of the Association for Computational Linguistics (Volume 2: Short Papers)}, pages 179--192, St. Julian{'}s, Malta. Association for Computational Linguistics.

\bibitem[{Omura et~al.(2024)Omura, Cheng, and Kurohashi}]{omura-etal-2024-empirical-study}
Kazumasa Omura, Fei Cheng, and Sadao Kurohashi. 2024.
\newblock \href {https://aclanthology.org/2024.lrec-main.96} {An empirical study of synthetic data generation for implicit discourse relation recognition}.
\newblock In \emph{Proceedings of the 2024 Joint International Conference on Computational Linguistics, Language Resources and Evaluation (LREC-COLING 2024)}, pages 1073--1085, Torino, Italia. ELRA and ICCL.

\bibitem[{Piedboeuf and Langlais(2025)}]{piedboeuf-langlais-2025-evaluation}
Fr{\'e}d{\'e}ric Piedboeuf and Philippe Langlais. 2025.
\newblock \href {https://aclanthology.org/2025.coling-main.231/} {On evaluation protocols for data augmentation in a limited data scenario}.
\newblock In \emph{Proceedings of the 31st International Conference on Computational Linguistics}, pages 3428--3443, Abu Dhabi, UAE. Association for Computational Linguistics.

\bibitem[{Prasad et~al.(2019)Prasad, Webber, Lee, and Joshi}]{AB2/SUU9CB_2019}
Rashmi Prasad, Bonnie Webber, Alan Lee, and Aravind Joshi. 2019.
\newblock \href {https://doi.org/11272.1/AB2/SUU9CB} {{Penn Discourse Treebank Version 3.0}}.

\bibitem[{Puri et~al.(2020)Puri, Spring, Shoeybi, Patwary, and Catanzaro}]{puri-etal-2020-training}
Raul Puri, Ryan Spring, Mohammad Shoeybi, Mostofa Patwary, and Bryan Catanzaro. 2020.
\newblock \href {https://doi.org/10.18653/v1/2020.emnlp-main.468} {Training question answering models from synthetic data}.
\newblock In \emph{Proceedings of the 2020 Conference on Empirical Methods in Natural Language Processing (EMNLP)}, pages 5811--5826, Online. Association for Computational Linguistics.

\bibitem[{Pyatkin et~al.(2023)Pyatkin, Yung, Scholman, Tsarfaty, Dagan, and Demberg}]{pyatkin-etal-2023-design}
Valentina Pyatkin, Frances Yung, Merel C.~J. Scholman, Reut Tsarfaty, Ido Dagan, and Vera Demberg. 2023.
\newblock \href {https://doi.org/10.1162/tacl_a_00586} {Design choices for crowdsourcing implicit discourse relations: Revealing the biases introduced by task design}.
\newblock \emph{Transactions of the Association for Computational Linguistics}, 11:1014--1032.

\bibitem[{Schick and Sch{\"u}tze(2021)}]{schick-schutze-2021-generating}
Timo Schick and Hinrich Sch{\"u}tze. 2021.
\newblock \href {https://doi.org/10.18653/v1/2021.emnlp-main.555} {Generating datasets with pretrained language models}.
\newblock In \emph{Proceedings of the 2021 Conference on Empirical Methods in Natural Language Processing}, pages 6943--6951, Online and Punta Cana, Dominican Republic. Association for Computational Linguistics.

\bibitem[{Scholman et~al.(2021)Scholman, Dong, Yung, and Demberg}]{scholman-etal-2021-comparison}
Merel Scholman, Tianai Dong, Frances Yung, and Vera Demberg. 2021.
\newblock \href {https://doi.org/10.18653/v1/2021.codi-main.9} {Comparison of methods for explicit discourse connective identification across various domains}.
\newblock In \emph{Proceedings of the 2nd Workshop on Computational Approaches to Discourse}, pages 95--106, Punta Cana, Dominican Republic and Online. Association for Computational Linguistics.

\bibitem[{Scholman et~al.(2022)Scholman, Dong, Yung, and Demberg}]{scholman2022DiscoGeM}
Merel C.~J. Scholman, Tianai Dong, Frances Yung, and Vera Demberg. 2022.
\newblock Discogem: A crowdsourced corpus of genre-mixed implicit discourse relations.
\newblock In \emph{Proceedings of the Thirteenth International Conference on Language Resources and Evaluation ({LREC}'22)}, Marseille, France. European Language Resources Association (ELRA).

\bibitem[{Sharma and Feldman(2023)}]{sharma2023team}
Ashwyn Sharma and David~I Feldman. 2023.
\newblock Team cadence at mediqa-sum 2023: Using chatgpt as a data augmentation tool for classifying clinical dialogue.
\newblock In \emph{CLEF (Working Notes)}, pages 1680--1687.

\bibitem[{Shi and Demberg(2019)}]{shi-demberg-2019-learning}
Wei Shi and Vera Demberg. 2019.
\newblock \href {https://doi.org/10.18653/v1/W19-0416} {Learning to explicitate connectives with {S}eq2{S}eq network for implicit discourse relation classification}.
\newblock In \emph{Proceedings of the 13th International Conference on Computational Semantics - Long Papers}, pages 188--199, Gothenburg, Sweden. Association for Computational Linguistics.

\bibitem[{Team~Gemma et~al.(2024)Team~Gemma, Riviere, Pathak, Sessa, Hardin, Bhupatiraju, Hussenot, Mesnard, Shahriari, Ram{\'e} et~al.}]{team2024gemma}
Team Team~Gemma, Morgane Riviere, Shreya Pathak, Pier~Giuseppe Sessa, Cassidy Hardin, Surya Bhupatiraju, L{\'e}onard Hussenot, Thomas Mesnard, Bobak Shahriari, Alexandre Ram{\'e}, et~al. 2024.
\newblock Gemma 2: Improving open language models at a practical size.
\newblock \emph{arXiv preprint arXiv:2408.00118}.

\bibitem[{Tiedemann(2012)}]{tiedemann-2012-parallel}
J{\"o}rg Tiedemann. 2012.
\newblock \href {http://www.lrec-conf.org/proceedings/lrec2012/pdf/463_Paper.pdf} {Parallel data, tools and interfaces in {OPUS}}.
\newblock In \emph{Proceedings of the Eighth International Conference on Language Resources and Evaluation ({LREC}'12)}, pages 2214--2218, Istanbul, Turkey. European Language Resources Association (ELRA).

\bibitem[{T{\"o}rnberg(2023)}]{tornberg2023chatgpt}
Petter T{\"o}rnberg. 2023.
\newblock Chatgpt-4 outperforms experts and crowd workers in annotating political twitter messages with zero-shot learning.
\newblock \emph{arXiv preprint arXiv:2304.06588}.

\bibitem[{Tzeng et~al.(2014)Tzeng, Hoffman, Zhang, Saenko, and Darrell}]{tzeng2014deep}
Eric Tzeng, Judy Hoffman, Ning Zhang, Kate Saenko, and Trevor Darrell. 2014.
\newblock Deep domain confusion: Maximizing for domain invariance.
\newblock \emph{arXiv preprint arXiv:1412.3474}.

\bibitem[{Ubani et~al.(2023)Ubani, Polat, and Nielsen}]{ubani2023zeroshotdataaug}
Solomon Ubani, Suleyman~Olcay Polat, and Rodney Nielsen. 2023.
\newblock Zeroshotdataaug: Generating and augmenting training data with chatgpt.
\newblock \emph{arXiv preprint arXiv:2304.14334}.

\bibitem[{Xiang et~al.(2022)Xiang, Wang, Dai, and Wang}]{xiang-etal-2022-connprompt}
Wei Xiang, Zhenglin Wang, Lu~Dai, and Bang Wang. 2022.
\newblock \href {https://aclanthology.org/2022.coling-1.75} {{C}onn{P}rompt: Connective-cloze prompt learning for implicit discourse relation recognition}.
\newblock In \emph{Proceedings of the 29th International Conference on Computational Linguistics}, pages 902--911, Gyeongju, Republic of Korea. International Committee on Computational Linguistics.

\bibitem[{Xiao et~al.(2020)Xiao, Huber, and Carenini}]{xiao-etal-2020-really}
Wen Xiao, Patrick Huber, and Giuseppe Carenini. 2020.
\newblock \href {https://doi.org/10.18653/v1/2020.codi-1.13} {Do we really need that many parameters in transformer for extractive summarization? discourse can help !}
\newblock In \emph{Proceedings of the First Workshop on Computational Approaches to Discourse}, pages 124--134, Online. Association for Computational Linguistics.

\bibitem[{Xue et~al.(2015)Xue, Ng, Pradhan, Prasad, Bryant, and Rutherford}]{xue-etal-2015-conll}
Nianwen Xue, Hwee~Tou Ng, Sameer Pradhan, Rashmi Prasad, Christopher Bryant, and Attapol Rutherford. 2015.
\newblock \href {https://doi.org/10.18653/v1/K15-2001} {The {C}o{NLL}-2015 shared task on shallow discourse parsing}.
\newblock In \emph{Proceedings of the Nineteenth Conference on Computational Natural Language Learning - Shared Task}, pages 1--16, Beijing, China. Association for Computational Linguistics.

\bibitem[{Yang et~al.(2024)Yang, Qian, Xu, Wang, and Xie}]{yang2024can}
Jianfei Yang, Hanjie Qian, Yuecong Xu, Kai Wang, and Lihua Xie. 2024.
\newblock Can we evaluate domain adaptation models without target-domain labels?
\newblock In \emph{The Twelfth International Conference on Learning Representations}.

\bibitem[{Yang et~al.(2020)Yang, Malaviya, Fernandez, Swayamdipta, Le~Bras, Wang, Bhagavatula, Choi, and Downey}]{yang-etal-2020-generative}
Yiben Yang, Chaitanya Malaviya, Jared Fernandez, Swabha Swayamdipta, Ronan Le~Bras, Ji-Ping Wang, Chandra Bhagavatula, Yejin Choi, and Doug Downey. 2020.
\newblock \href {https://doi.org/10.18653/v1/2020.findings-emnlp.90} {Generative data augmentation for commonsense reasoning}.
\newblock In \emph{Findings of the Association for Computational Linguistics: EMNLP 2020}, pages 1008--1025, Online. Association for Computational Linguistics.

\bibitem[{Yarowsky(1995)}]{yarowsky1995unsupervised}
David Yarowsky. 1995.
\newblock \href {https://aclanthology.org/P95-1026} {Unsupervised word sense disambiguation rivaling supervised methods}.
\newblock In \emph{33rd annual meeting of the association for computational linguistics}, pages 189--196.

\bibitem[{Yung et~al.(2024)Yung, Ahmad, Scholman, and Demberg}]{yung-etal-2024-prompting}
Frances Yung, Mansoor Ahmad, Merel Scholman, and Vera Demberg. 2024.
\newblock \href {https://aclanthology.org/2024.law-1.15} {Prompting implicit discourse relation annotation}.
\newblock In \emph{Proceedings of The 18th Linguistic Annotation Workshop (LAW-XVIII)}, pages 150--165, St. Julians, Malta. Association for Computational Linguistics.

\bibitem[{Yung et~al.(2022)Yung, Anuranjana, Scholman, and Demberg}]{yung-etal-2022-label}
Frances Yung, Kaveri Anuranjana, Merel Scholman, and Vera Demberg. 2022.
\newblock \href {https://aclanthology.org/2022.codi-1.7} {Label distributions help implicit discourse relation classification}.
\newblock In \emph{Proceedings of the 3rd Workshop on Computational Approaches to Discourse}, pages 48--53, Gyeongju, Republic of Korea and Online. International Conference on Computational Linguistics.

\bibitem[{Yung and Demberg(2025)}]{yung-demberg-2025-crowdsourcing}
Frances Yung and Vera Demberg. 2025.
\newblock \href {https://aclanthology.org/2025.comedi-1.2/} {On crowdsourcing task design for discourse relation annotation}.
\newblock In \emph{Proceedings of Context and Meaning: Navigating Disagreements in NLP Annotation}, pages 12--19, Abu Dhabi, UAE. International Committee on Computational Linguistics.

\bibitem[{Zeng et~al.(2024)Zeng, He, Sun, Xu, Liu, and Wang}]{zeng-etal-2024-global}
Lei Zeng, Ruifang He, Haowen Sun, Jing Xu, Chang Liu, and Bo~Wang. 2024.
\newblock \href {https://aclanthology.org/2024.lrec-main.686} {Global and local hierarchical prompt tuning framework for multi-level implicit discourse relation recognition}.
\newblock In \emph{Proceedings of the 2024 Joint International Conference on Computational Linguistics, Language Resources and Evaluation (LREC-COLING 2024)}, pages 7760--7773, Torino, Italia. ELRA and ICCL.

\bibitem[{Zhao et~al.(2023)Zhao, He, Xiao, and Xu}]{zhao-etal-2023-infusing}
Haodong Zhao, Ruifang He, Mengnan Xiao, and Jing Xu. 2023.
\newblock \href {https://doi.org/10.18653/v1/2023.acl-long.357} {Infusing hierarchical guidance into prompt tuning: A parameter-efficient framework for multi-level implicit discourse relation recognition}.
\newblock In \emph{Proceedings of the 61st Annual Meeting of the Association for Computational Linguistics (Volume 1: Long Papers)}, pages 6477--6492, Toronto, Canada. Association for Computational Linguistics.

\bibitem[{Zhou et~al.(2022)Zhou, Lan, Wu, Chen, and Ma}]{zhou-etal-2022-prompt-based}
Hao Zhou, Man Lan, Yuanbin Wu, Yuefeng Chen, and Meirong Ma. 2022.
\newblock \href {https://doi.org/10.18653/v1/2022.findings-emnlp.282} {Prompt-based connective prediction method for fine-grained implicit discourse relation recognition}.
\newblock In \emph{Findings of the Association for Computational Linguistics: EMNLP 2022}, pages 3848--3858, Abu Dhabi, United Arab Emirates. Association for Computational Linguistics.

\bibitem[{Zhou et~al.(2020)Zhou, Yang, Hospedales, and Xiang}]{zhou2020deep}
Kaiyang Zhou, Yongxin Yang, Timothy Hospedales, and Tao Xiang. 2020.
\newblock Deep domain-adversarial image generation for domain generalisation.
\newblock In \emph{Proceedings of the AAAI conference on artificial intelligence}, volume~34, pages 13025--13032.

\end{thebibliography}

\appendix

%\clearpage

\section{Data}\label{sec:data}

\begin{table}[ht]
\small \centering
  \begin{tabular}{@{}l@{}|ll@{}|lll@{}|l@{}}
         \hline
         & \multicolumn{2}{c|}{PDTB 3.0}     
         & \multicolumn{4}{c}{DG 1.5 (test)} \\
         \cline{2-7}
         & train & dev & EP & WK & NV & ttl.\\

         \hline\hline
         \textbf{Expansion}&&&&&\\
\textsc{conjunc.}&3584&298&314&177&323&814\\
\textsc{lev.-of-det.}&2493&262&532&161&460&1153\\
\textsc{instant.}&1117&116&212&37&94&343\\
\textsc{manner}&191&14&20&0&6&26\\
\textsc{substitut.}&278&27&41&4&47&92\\
\textsc{equival.}&252&25&50&2&38&90\\
\textsc{disjunct.}&0&0&0&0&5&5\\
\textsc{exception}&0&0&0&1&6&10\\

        \hline \textbf{Contingency}&&&&&\\
\textsc{cause}&4469&450&885&86&857&1828\\
\textsc{purpose}&1102&97&139&10&49&198\\
\textsc{cause+bel.}&157&13&0&0&0&0\\
\textsc{condition}&152&18&85&3&19&107\\

        \hline \textbf{Contrast} &&&&&\\
\textsc{concession}&1164&103&229&23&186&438\\
\textsc{contrast}&639&82&38&19&58&115\\
\textsc{similarity}&0&0&65&6&42&113\\

        \hline \textbf{Temporal}&&&&&\\
\textsc{asynchr.}&985&102&18&70&536&624\\
\textsc{synchr.}&433&34&73&16&158&247\\

\hline\hline
\textbf{Total}&17016 & 1641 &2704 &615&2884&6203 \\

           \hline
    \end{tabular}

   \caption{Distribution of the Level-2 classes, grouped under 4 Level-1 categories, in each data subset. DiscoGeM's distribution is based on the single majority label per sample
   \label{tab:dist}}
    \end{table}

We use PDTB~3.0 \cite{AB2/SUU9CB_2019} as the source-domain data for training and tuning and DiscoGeM~1.5 \cite{scholman2022DiscoGeM,yung-demberg-2025-crowdsourcing} as the target-domain data for evaluation.
Table~\ref{tab:dist} shows the distribution of all labelled data used in this study. 
 
PDTB~3.0 is the largest discourse resource in English annotated by trained annotators.  The texts come from the news articles of the Wall Street Journal in the 90s. Implicit relations are annotated between consecutive sentences as well as within individual sentences, if identified.  The relation labels are arranged in a 3-level hierarchy.

We train our source-domain model to predict the 14 Level-2 labels with more than $10$ instances in the test set, as in previous works \cite[e.g.][]{kim-etal-2020-implicit}. Sections $2-20$ and $1-2$ are used as for training and tuning respectively \cite{ji-etal-2015-closing}.

DiscoGeM~1.5 is a crowdsourced corpus of implicit discourse relations in English containing texts from multiple genres: European Parliament proceedings (EP), Wikipedia articles (WK), and literature (NV).

Each relation is annotated by $10$ crowdworkers using a connective insertion task. The label set is also based on the PDTB 3.0 label hierarchy, but only inter-sentential relations are annotated. We use the complete DG corpus for evaluation, except the instances that are labelled \textsc{no relation}, which is not considered as a type of implicit DRs in PDTB~3.0.

In the training and tuning of all the models, we use a single label per instance (\textit{Conn1SenseClass1} label of PDTB). The predicted labels are evaluated against multiple labels, which are defined as annotations with $40\%$ or more votes. 

\section{Prompt template} \label{sec:prompt}

The exact DC-prompt and the DR-prompt templates are shown in Figure~\ref{fig:prompt}. Table~\ref{tab:conn_map} lists the connectives used in the DC-prompt for each DR label.

\begin{figure} [ht]
\framedtext{\small
\#\#\#Instructions\#\#\#\\
Complete the sentence, and don't generate more than one sentence.\\

\#\#\#Example\#\#\#\\
Q:   \textcolor{black}{The Artist has his routine. He spends his days sketching passers-by, or trying to.} \textcolor{black}{Later,} ...\\
A:  \textcolor{black}{at night he returns to the condemned building he calls home.}\\

\#\#\#Your task\#\#\#\\
Q:  \textcolor{black}{The brokerage firms learned a lesson the last time around.  Therefore, ...}\\
A:
}

\framedtext{\small
\#\#\#Instructions\#\#\#\\
Given two arguments, the relation \textcolor{black}{"Conjunction"} is defined as \textcolor{black}{"both arguments, which don’t directly relate to each other, bear the same relation to some other situation evoked in the discourse".}

Here are examples that have the relation \textcolor{black}{"Conjunction"}:\\
She, out of gratitude, had her arms wrapped around his neck as they slept. CONJUNCTION\\

Various articles of their clothing lay intermingled around the bed.\\

\#\#\#Your task\#\#\#\\
Please write down the second arguments that have the relation CONJUNCTION to the first argument: "And over the desert plain one heard only the moan of squalls through the broken trellises of the enclosures." Here list several second arguments:
}
\caption{Top: DC-prompt; bottom: DR-Prompt}
\label{fig:prompt}
\end{figure}

\begin{table}[h]
    \centering \small
    \begin{tabular}{@{}l|l@{}}
        \hline
         intended DR $L'$ & DC in DC-prompt  \\
         \hline
         \textsc{conjunction} & In addition, | Furthermore, \\
         \textsc{level-of-detail} & More specifically, | In particular, \\
         \textsc{instantiation} & For example, | For instance, \\
         \textsc{manner} & by | by means of \\
         \textsc{substitution} & Instead, | Rather than that, \\
         \textsc{equivalence} & In other words, | That is to say, \\
         \textsc{cause} & It is/was because | Therefore, \\
         \textsc{purpose} & in order | so as\\
         \textsc{cause+belief} & As an evidence, | This justifies that \\
         \textsc{condition} & if | if it is/was \\
         \textsc{concession} & Nonetheless, | Nevertheless,\\
         \textsc{contrast} & On the other hand, | In contrast, \\
         \textsc{asynchronous} & Later, |Subsequently, \\
         \textsc{synchronous} & Simultaneously, | Meanwhile, \\ 
         \hline
    \end{tabular}
    \caption{The discourse connectives used in the DC-prompt for different DR types. 
}
    \label{tab:conn_map}
\end{table}

%%%%%%%%%%%%%%%%%%%%%%%%%%%%%%%%%%%%%%%%%%%%%%%%%%%%%%%%%%%%%%%%

\section{Screening methods} \label{sec:confusion screen}

The \textit{confusion screen} filters out samples where the base model's predicted label does not match the intended label but instead corresponds to a frequent misclassification of the ground truth. For example, if the baseline model frequently misclassifies \textsc{cause+belief} as \textsc{cause}, synthetic samples labeled as \textsc{cause+belief} but predicted as \textsc{cause} are excluded by the confusion screen. Table~\ref{tab:l_confuse} provides a complete mapping of the most common mispredictions, derived from the confusion matrix of the \textit{RoBERTa-base} model evaluated on the PDTB~3.0 dev set. This screening method follows the strategy proposed in previous work \cite{omura-etal-2024-empirical-study}, but replaces zero-shot prompting-based predictions with those obtained through supervised classification.
\begin{table}[htpb]
    \centering \small
    \begin{tabular}{@{}l|l@{}}
        \hline
        intended label $L'$ & confuse$(L')$ \\
        \hline\hline
        \textsc{conjuntion},\textsc{level-of-detail} &\textsc{cause}\\
        \textsc{substitution},  \textsc{equivalence} &\\
        \textsc{cause+belief}, \textsc{condition}&\\
        \textsc{concession}, \textsc{asynchronous} &\\
        \hline
        \textsc{instantiation}, \textsc{manner}, \textsc{cause} & \textsc{level-of-detail}\\
        \hline
        \textsc{synchronous}, \textsc{similarity}& \textsc{conjunction}\\
        
        \hline
        \textsc{purpose} & \textsc{condition}\\
        \hline
        \textsc{contrast} & \textsc{concession}\\        
        \hline
    \end{tabular}
    \caption{Intended label $L'$ vs confuse$(L')$ used in the \textbf{confuse screen}. The generation is selected if $L_{pred} \neq$ confuse$(L')$. }
    \label{tab:l_confuse}
\end{table}
Table \ref{tab:generation} summarizes the screened label distributions per different settings.

The screened data size of the \textit{mistral} generation is $10\%-20\%$ larger, indicating higher agreement with the supervised model. On the other hand, the high selection rate of the confusion screen used in the previous work suggests a significantly more lenient selection process.
\begin{table}[h]
\centering \small
  \begin{tabular}{l|l|l|l}
\hline
\textbf{LLM}&llama3&gemma2& mistral\\
\hline
\textbf{prompt}&\multicolumn{3}{c}{DC}\\
\hline
\textbf{screen}&\multicolumn{3}{c}{strict}\\
\hline
\textbf{EP}&9361&8464&10724\\
\textbf{WK}&9101&9068&10689\\
\textbf{NV}&7579&7415&10224\\
\hline\hline
\textbf{LLM}&\multicolumn{3}{c}{mistral}\\
\hline
\textbf{prompt}&DC&DC&DR\\
\hline
\textbf{screen}&confuse&smooth&strict\\
\hline
\textbf{EP}&39846&17337&11473\\
\textbf{WK}&44833&19434&12457\\
\textbf{NV}&44962&18087&13198\\
 \hline
    \end{tabular}
\caption{Size of the synthetic data generated by different LLMs, prompts, and screens. There were $60000$ instances ($4000$ $Arg1_{\text{raw}}$s $\times$ $15$ DR types) generated in each case before screening. }
\label{tab:generation}
\end{table}
%
%
%%%%%%%%%%%%%%%%%%%%%%%%%
\clearpage
\section{Annotated and synthetic DR examples} \label{sec:examples}

 \begin{figure} [htpb]
\small
 1a)
\framedtext{
Arg1:It is an honour and a pleasure to have the opportunity to present this report to Parliament today.	\\
Arg2: It is on the very important subject of product liability on which the European Community first introduced legislation as long ago as 1985 in the form of a directive...
}
Annotated \textsc{reason} relation\\

1b)
\framedtext{
Arg1: Mrs Beer asked me to pass on her thanks to the various political group coordinators and to the Council and Commission for their good cooperation in the fruitful compromise negotiations. \\
Arg2: The fruitful compromise negotiations resulted in a mutually beneficial agreement, as evidenced by the expressed gratitude from Mrs Beer. 
}
Synthetic \textsc{reason} relation\\

2a)
\framedtext{
Arg1: The history of agriculture began thousands of years ago.	
Arg2: After gathering wild grains beginning at least 105,000 years ago, nascent farmers began to plant them around 11,500 years ago. 
}
Annotated \textsc{arg2-as-instant} relation\\

2b)
\framedtext{
Arg1: Holmes mourned that the pony pennings of his day were only "a shadow of their former glory".\\
Arg2: Breeds such as Shire horses or Friesians, once prominent in England and the Netherlands respectively, could serve as examples.
}
Synthetic \textsc{arg2-as-instant} relation\\

\caption{Examples taken from DiscoGeM~1.5 and the screened synthetic samples generated by \textit{Mistral}. }
\label{fig:examples}
\end{figure}

\end{document}